\documentclass[sigconf, screen, nonacm]{acmart}

\usepackage{graphicx}
\usepackage{caption}
\usepackage{float}
\usepackage{multirow}
\usepackage{booktabs}
\usepackage{multicol}
\usepackage{amsmath}
\usepackage[capitalize]{cleveref}
\usepackage{xcolor}

\usepackage{color}

\AtBeginDocument{%
  }

\copyrightyear{2025}
\acmYear{2025}
\setcopyright{cc}
\setcctype{by-nc-sa}
\acmConference[MM '25]{Proceedings of the 33rd ACM International Conference on Multimedia}{October 27--31, 2025}{Dublin, Ireland}
\acmBooktitle{Proceedings of the 33rd ACM International Conference on Multimedia (MM '25), October 27--31, 2025, Dublin, Ireland}\acmDOI{10.1145/3746027.3754718}
\acmISBN{979-8-4007-2035-2/2025/10}

\begin{document}

\title{Gather and Trace: Rethinking Video TextVQA from an Instance-oriented Perspective}


\author{Yan Zhang}
\affiliation{%
  \institution{Institute of Information Engineering, Chinese Academy of Sciences}
  \institution{School of Cyber Security, University of Chinese Academy of Sciences}
  \city{Beijing}
  \country{China}
}
\email{zhangyan2022@iie.ac.cn}

\author{Gangyan Zeng}
\affiliation{%
  \institution{School of Cyber Science and Engineering, Nanjing University of Science and Technology}
  \city{Nanjing}
  \country{China}
}
\email{gyzeng@njust.edu.cn}

\author{Daiqing Wu}
\affiliation{%
  \institution{Institute of Information Engineering, Chinese Academy of Sciences}
  \institution{School of Cyber Security, University of Chinese Academy of Sciences}
  \city{Beijing}
  \country{China}
}
\email{wudaiqing@iie.ac.cn}

\author{Huawen Shen}
\affiliation{%
  \institution{Institute of Information Engineering, Chinese Academy of Sciences}
  \institution{School of Cyber Security, University of Chinese Academy of Sciences}
  \city{Beijing}
  \country{China}
}
\email{shenhuawen@iie.ac.cn}

\author{Binbin Li}
\authornote{Corresponding author}
\affiliation{%
  \institution{Institute of Information Engineering, Chinese Academy of Sciences}
  \institution{School of Cyber Security, University of Chinese Academy of Sciences}
  \city{Beijing}
  \country{China}
}
\email{libinbin@iie.ac.cn}

\author{Yu Zhou}
\affiliation{%
  \institution{VCIP \& TMCC \& DISSec, College of Computer Science, Nankai University} 
  \city{Tianjin}
  \country{China}
}
\email{yzhou@nankai.edu.cn}

\author{Can Ma}
\authornote{Corresponding author}
\affiliation{%
  \institution{Institute of Information Engineering, Chinese Academy of Sciences}
  \institution{School of Cyber Security, University of Chinese Academy of Sciences}
  \city{Beijing}
  \country{China}
}
\email{macan@iie.ac.cn}

\author{Xiaojun Bi}
\affiliation{%
  \institution{Key Laboratory of Ethnic Language Intelligent Analysis and Security Governance of MOE, Minzu University of China}
  \city{Beijing}
  \country{China}
}
\email{bixiaojun@hrbeu.edu.cn}

\renewcommand{\shortauthors}{Yan Zhang et al.}

\begin{abstract}
  Video text-based visual question answering (Video TextVQA) aims to answer questions by explicitly reading and reasoning about the text involved in a video. Most works in this field follow a frame-level framework which suffers from redundant text entities and implicit relation modeling, resulting in limitations in both accuracy and efficiency. In this paper, we rethink the Video TextVQA task from an instance-oriented perspective and propose a novel model termed GAT (Gather and Trace). First, to obtain accurate reading result for each video text instance, a context-aggregated instance gathering module is designed to integrate the visual appearance, layout characteristics, and textual contents of the related entities into a unified textual representation. Then, to capture dynamic evolution of text in the video flow, an instance-focused trajectory tracing module is utilized to establish spatio-temporal relationships between instances and infer the final answer. Extensive experiments on several public Video TextVQA datasets validate the effectiveness and generalization of our framework. GAT outperforms existing Video TextVQA methods, video-language pretraining methods, and video large language models in both accuracy and inference speed. Notably, GAT surpasses the previous state-of-the-art Video TextVQA methods by 3.86\% in accuracy and achieves ten times of faster inference speed than video large language models. The source code is available at https://github.com/zhangyan-ucas/GAT.

  
\end{abstract}





\maketitle

\section{Introduction}
\begin{figure*}[htbp]
  \centering
    \includegraphics[width=0.8\textwidth]{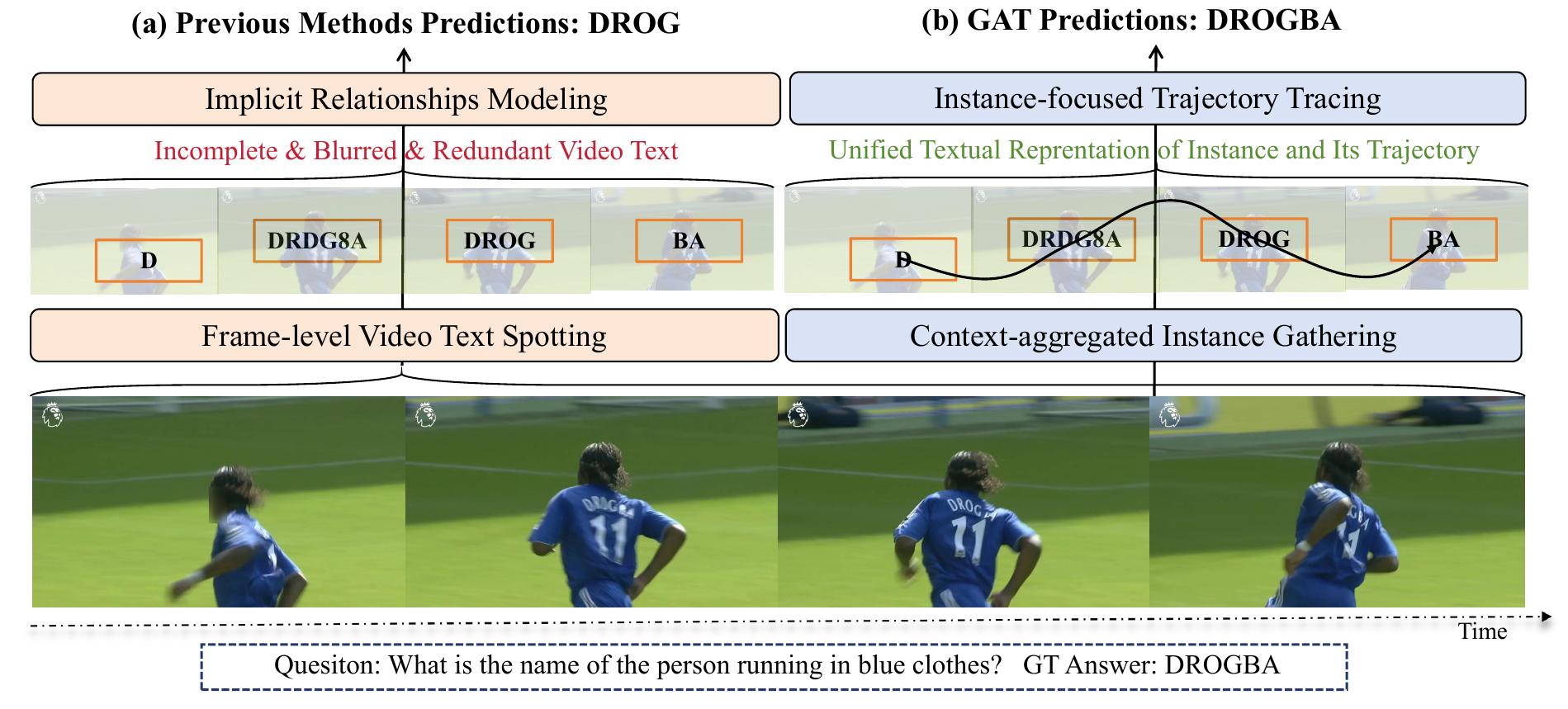}
  \caption{A representative example to illustrate the difference between previous frame-level methods and our instance-oriented method. (a) Previous methods rely on frame-level Video Text Spotting (VTS) techniques to extract text and implicit spatio-temporal modeling to infer answers, thus suffering from numerous low-quality text entities and confusing reasoning processes. (b) The proposed GAT method develops a context-aggregated instance gathering module to encode a unified textual representation for each instance, as well as an instance-focused trajectory tracing module to recover the dynamic evolution of video text, enhancing performance in both accuracy and efficiency.}
  \label{fig:figure1}
\end{figure*}

Understanding video text is crucial for interpreting the physical world, driving numerous applications including intelligent security monitoring \cite{rekha2025intelligent}, autonomous driving systems \cite{zhang2021character}, and scene text manipulation \cite{zeng2024textctrl,shu2025visual,zeng2024focus,shen2025ldp}. Against this backdrop, the research community has introduced a video text understanding task known as video text-based visual question answering (Video TextVQA) \cite{tea, M4-ViteVQA}. Compared to common video question answering (VideoQA) \cite{videoqa}, this task is more challenging as it explicitly requires models to read and comprehend video text that appears in each video frame to answer the given questions.

Recently, numerous Video TextVQA methods have achieved remarkable progress by adopting a two-stage paradigm \cite{M4-ViteVQA, tea}. In this paradigm, off-the-shelf Video Text Spotting (VTS) techniques \cite{cotext,gloma,gomatching,transdetr,eve,gloma} are first utilized to extract video text from each video frame. After modeling the spatio-temporal relationships of these extracted results, generative models like T5 \cite{t5} are responsible for predicting answers in an auto-regressive manner. For instance, the pioneering Video TextVQA method, T5-ViteVQA \cite{M4-ViteVQA}, introduces five Transformer blocks to encode modality-specific features across video frames, and then feeds them into a T5-like Transformer encoder-decoder to infer the answer. Subsequently, TEA \cite{tea} establishes a simple yet effective baseline by mining the spatio-temporal clues among visual entities across multiple frames and performs reasoning through video text-aware aggregation.


Despite the promising results, existing approaches are essentially modeling at the frame level (as shown in \cref{fig:figure1} (a)), leading to limitations in terms of accuracy and efficiency. 
\textbf{During the VTS stage,} due to the dynamic nature of video text, there are a lot of low-quality (incomplete, blurred, or reflected) text entities in video frames. If their spotted results are taken directly as the downstream input, the model will be subject to huge noise interference and error accumulation. Our analysis of the commonly used VTS dataset ICDAR15 \cite{icdar15} reveals that about 65\% of text instances exhibit varying visual characteristics across video frames, hinting that many of the spotted results are in fact unreliable.
\textbf{During the reasoning stage,} 
following general VideoQA’s practice \cite{videoqa}, current efforts tend to employ absolute-location embedding and temporal index to implicitly represent spatial and temporal information, respectively.
As the dynamic evolution of video text is not well captured, this scheme cannot distinguish redundant text contents and associate their intricate relationships, thus impairing the performance and interpretability of the model. In addition, even with a frame-level sampling strategy commonly adopted in the field of general video understanding \cite{yi2024video, liu2024ppllava, yuan2025tarsier2}, overly lengthy inputs will still affect the reasoning speed, thus hindering the application in real scenarios.



Based on the above analysis, we think about how to break through the bottleneck of the frame-level framework. Actually, when understanding text in videos, human beings often treat video text instances as fundamental units, which determines text contents by associating the context in consecutive frames and locating the dynamic motion trajectories when building relationships. 

Inspired by this, we resort to address the Video TextVQA task from an instance-oriented perspective.  To be specific, the concept of instance refers to a continuous occurrence of video text within a video, even if it undergoes visual changes across frames. For example, in \cref{fig:figure1}, ``DRDG8A'' in the second frame and ``DROG'' in the third frame are defined as the same instance.

Following the spirit of instance-oriented Video TextVQA, a model termed GAT (Gather and Trace) is proposed. As illustrated in \cref{fig:figure1} (b), GAT improves the Video TextVQA performance from two aspects. 
\textbf{(1) Context-aggregated Instance Gathering.} To avoid the effects of spotting errors caused by low-quality video text, we propose to gather related entities and derive a comprehensive result for each video text instance. Building upon the VTS method \cite{gomatching}, GAT integrates the rich context information, including visual appearance, layout characteristics, and textual contents of the text instance throughout the video into a unified representation. Meanwhile, a video text recognition loss and a video text-aware auxiliary loss are utilized to determine whether the video text instance that appears in the current frame is unclear or incomplete.
\textbf{(2) Instance-focused Trajectory Tracing.} Unlike previous methods constrained by redundant entity modeling, we focus on each unique text instance and explicitly construct the corresponding trajectory. To achieve this, an instance trajectory-aware attention mechanism is designed, in which trajectory distance is defined to measure the relative relations between instances and then used to calibrate the attention in cross-instance interactions. Following it, a Transformer-based decoder predicts answers via reasoning from the trajectory clues and locating the critical video text.


Through extensive experiments on several public datasets \cite{ M4-ViteVQA, roadtextvqa}, the effectiveness of our proposed framework is demonstrated. GAT outperforms existing Video TextVQA methods, video-language pretraining methods, and video large language models (Video-LLMs) in both accuracy and inference speed. Besides, thanks to saving on processing redundant input tokens, GAT exhibits obvious advantages in model complexity and inference speed.

The contribution of this work can be summarized as:
\begin{itemize}
\item {After rethinking the conventional Video TextVQA methods that are bounded by the frame-level framework in terms of accuracy and efficiency, we propose the instance-oriented perspective for reading and comprehending video text.}

\item {To obtain accurate reading results from numerous interfering entities, we develop a context-aggregated instance gathering module that aggregates the visual appearance, layout characteristics, and textual contents of each instance across the video sequence. To infer reasonable answers within the dynamic video flow, we utilize an instance-focused trajectory tracing module to ensure the completeness of each text trajectory and perform trajectory-aware instance interactions with the questions.}

\item {Extensive experiments verify the effectiveness and superiority of our proposed method on several representative datasets. The results indicate that GAT reaches a new state-of-the-art on the Video TextVQA task.}

\end{itemize}

\section{Related Work}

\subsection{Image TextVQA}
Image text-based visual question answering (referred to as Image TextVQA) involves answering questions based on reading and understanding textual content within a single image. Generally, image TextVQA approaches \cite{SETS, biten2019scene, mishra2019ocr,  wang2020general, zhu2021simple, xenos2023simple, zhou2025egotextvqa, zhou2023exploring, zhou2024graph, zeng2023beyond, shen2023divide} predominantly follow a two-stage paradigm. In this paradigm, a lightweight OCR system \cite{du2025instruction,ipad,zhang2025linguistics,zheng2024cdistnet,lyu2025textblockv2,lyu2025arbitrary,wang2022tpsnet} initially extracts scene text from images, which is then fed into a Transformer-based multi-modal fusion framework to predict answers. As an initial attempt, Singh \textit{et al.} \cite{singh2019towards} first conduct single hop attention conditioned on the given question, and then predict answers from fixed vocabularies or OCR tokens. M4C \cite{m4c} leverages a Transformer-based multimodal architecture and decodes answers auto-regressively with a dynamic pointer network. Building on top of M4C, SA-M4C \cite{sa-m4c} considers twelve spatial relations between visual entities to assist self-attention in the multimodal Transformer. With the advent of visual-language pretraining techniques, TAP \cite{yang2021tap} introduces Image TextVQA-specific pretraining tasks to better align visual entities, including scene text and visual objects, thereby enhancing scene text-aware image understanding. 

More recently, generative paradigms have emerged, with Image TextVQA methods \cite{t5, latr, prestu, sal, fitb} harnessing the reasoning capabilities and extensive common sense knowledge of language models to improve performance on visual question answering tasks. Specifically, LaTr \cite{latr} utilizes T5 \cite{t5} as the reasoning backbone and addresses the question-answering problem in a text-generation way. Similarly to LaTr, PreSTU \cite{prestu} proposes the SPLITOCR pretraining objective to learn to read text in scene images. Furthermore, FITB \cite{fitb} proposes a text-centered generative framework in which multimodal information is represented mainly in textual form and rationale-augmented prompting is involved.

\subsection{Video TextVQA}

To better meet the requirements in real-world scenarios, Video TextVQA aims to answer text-related questions according to a given video, which brings extra challenges \cite{M4-ViteVQA, zhou2025egotextvqa, roadtextvqa, yang2025vidtext}. Unlike Image TextVQA, this task is considerably more complex due to the dynamic nature of videos, where answers may appear in arbitrary video frames. An intuitive solution to the challenge is extending Image TextVQA methods to the video domain. Typically, a two-stage paradigm is adopted: frozen VTS techniques are first utilized to read video text, based on which the answers are predicted in an auto-regressive manner. Specifically, T5-ViteVQA \cite{M4-ViteVQA} employs a frame sampling strategy widely adopted in the video understanding field. It samples multiple frames uniformly and extracts video text, visual features, and temporal features within them. These features are subsequently fed into a T5-like Transformer encoder-decoder to infer the answer. More recently, TEA \cite{tea} further boosts performance by exploiting spatio-temporal information among video text and highlighting the role of questions with an aggregation module.

Nevertheless, existing work mostly follows a frame-level framework that deviates from the manner in which humans recognize video text. Considering that it is enough to understand video text by modeling the complete content and motion trajectory of text instances, we explore addressing the Video TextVQA task from an instance-oriented perspective.

\begin{figure*}[t]
    \centering  
 
  \includegraphics[width=0.83\textwidth]{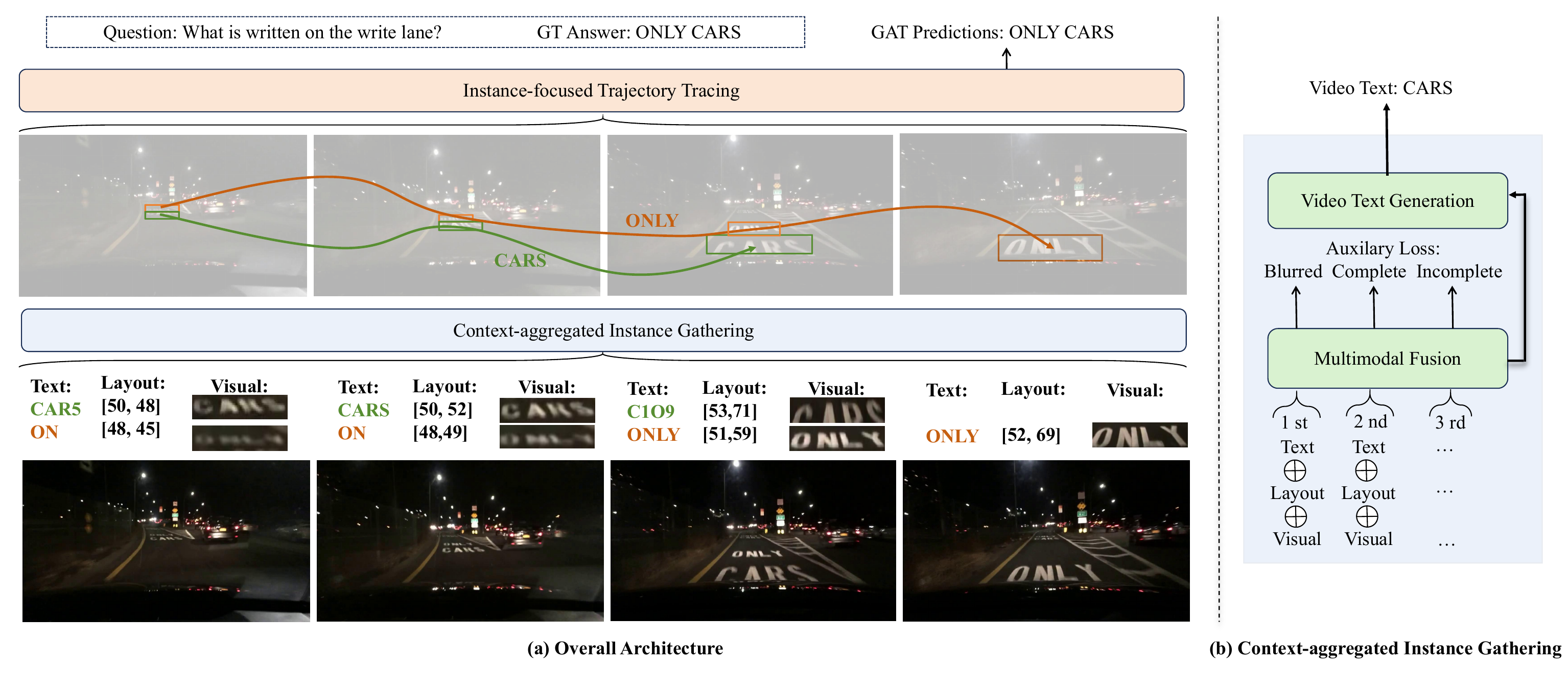}

   \caption{
  (a) Overall architecture of GAT. (b) Illustration of the context-aggregated instance gathering module. 
}
    \label{fig:figure2}
\end{figure*}

\section{Method}

 Considering the presence of video text with visual variations across dynamic video sequences, such as incomplete, blurred, and reflected appearances, we develop an instance-oriented framework. In this framework, each video text instance is treated as the fundamental unit, which refers to the continuous occurrence of video text within a video, even if it undergoes visual changes across frames. For instance, in \cref{fig:figure2}, ``CAR5" in the first frame and ``CARS" in the second frame are regarded as the same instance, while ``C1O9" and ``ONLY" in the third frame are identified as distinct instances. 

An overview of the proposed GAT framework is illustrated in \cref{fig:figure2}. GAT adopts a two-stage paradigm where multiple instances are first extracted and then fed into a generative model \cite{t5} to generate answers in an auto-regressive manner. Building upon the VTS method \cite{gomatching}, in \cref{section:3.1}, the context-aggregated instance gathering module specializes in localizing, identifying, and tracking multiple instances across dynamic video sequences. Following it, the instance-focused trajectory tracing module in \cref{section:3.2} leverages a trajectory-aware attention mechanism to facilitate instance-level interactions and utilizes a Transformer decoder to infer the final answer.

\subsection{Context-aggregated Instance Gathering}
\label{section:3.1}
We note that the Video TextVQA task emphasizes the global spatio-temporal dynamics of instances across entire video sequences, whereas the VTS task focuses on the visual appearance of the instance within individual frames. Previous Video TextVQA methods \cite{tea, M4-ViteVQA} employ a frame sampling strategy where separate frames are sampled and video text within these frames is fed into the generative model. However, this strategy compromises the global spatio-temporal information of instances and leads to suboptimal performance in video understanding tasks. As illustrated in \cref{fig:figure2} (b), building upon the existing VTS method \cite{gomatching}, we aggregate global spatio-temporal dynamics for each video text instance to bridge the gap between the VTS task and the Video TextVQA task. Specifically, we feed the multiple frame-level results for a single video text instance extracted via the VTS method into a lightweight encoder-decoder Transformer architecture to generate a unified textual representation of the instance.

\subsubsection{Instance-oriented Multimodal Features in Video Sequences} Video text instances involved in the video flow naturally encompass rich multimodal context features. To this end, we utilize the existing VTS model GoMatching \cite{gomatching} to extract text entities throughout the video sequence and integrate them into unique instances for processing. Specifically, each instance in a video frame can be characterized by three multimodal features: (1) the textual content encoded via a token embedding layer; (2) the layout characteristics represented through the normalized bounding box coordinates of that instance; (3) the visual appearance extracted by ROI pooling \cite{deepsolo} over the instance's bounding box. Furthermore, GoMatching \cite{gomatching} is capable of tracking all multimodal features of a video text instance throughout the video. We introduce a set of dedicated tokens to connect multimodal features from entire videos into a coherent sequence. In \cref{sec: section3.1.2}, we feed these multimodal features into a lightweight Transformer encoder-decoder architecture to predict the unified representations of video text instances.

We deem that a unified representation of the video text instance should contain the most complete visual content across the video and its trajectory. Therefore, by considering these multimodal features, our approach confers three key benefits: (1) As the basic VTS method is pretrained on extensive image spotting data, its recognition results in video frames are generally reliable, providing a solid reference for predicting global representation for the video text instance. (2) The coordinates of a video text instance in the current frame indicate whether it enters or exits the video. Video text instances entering or exiting the video often have incomplete visual presentations. Therefore, their recognition results should not be referenced. (3) The visual appearance of the video text instance can be an indicator to reflect whether the instance in the current frame is blurred or clear. Following this, we can disregard the recognition results from unclear frames.

\subsubsection{Optimization for context-aggregated instance gathering}
\label{sec: section3.1.2}

In our design, we expect to aggregate the unified representation of each video text instance, which encompasses the most complete visual content across the video and its trajectory. For common VTS datasets such as ICDAR15 \cite{icdar15}, the annotations are only labeled for the currently visible text without the consideration of its temporal evolution characteristics, so they may be incomplete. To address this, we redefine video text instances automatically, ensuring that video text is annotated based on its most complete and legible appearance in the video.

Concretely, we feed multimodal features of video text instances across multiple frames to a lightweight Transformer-based encoder-decoder architecture to aggregate the unified representation. Within the encoder, these features interact to identify which frame's features are most useful for generating the unified representation. We introduce an auxiliary loss term to determine which original recognition results align with the annotated ground truth. Additionally, we predict the unified representation of the video text instance in an auto-regressive way. The training objective for context-aggregated instance gathering is formulated in \cref{equ:equation1}. 


\begin{equation}
\label{equ:equation1}
\begin{cases}
    \displaystyle \mathcal{L}_{rec} = - \frac{1}{N}\sum_{t=1}^{N}\log(p_{t}\mid g_{t}),\\
    \mathcal{L}_{aux} = -\sum_{k=1}^{S} y_k \log(\hat{y}_k),\\
    \mathcal{L}_{perc} = \mathcal{L}_{rec} + \lambda \cdot \mathcal{L}_{aux},
\end{cases}
\end{equation}
where we use the cross-entropy loss for video text recognition. pt and gt are the prediction and ground truth at time step t. $\lambda$ is a hyperparameter to balance the two losses.

\subsection{Instance-focused Trajectory Tracing}
\label{section:3.2}
\begin{figure}[t]
    \centering  
  \includegraphics[width=0.30\textwidth]{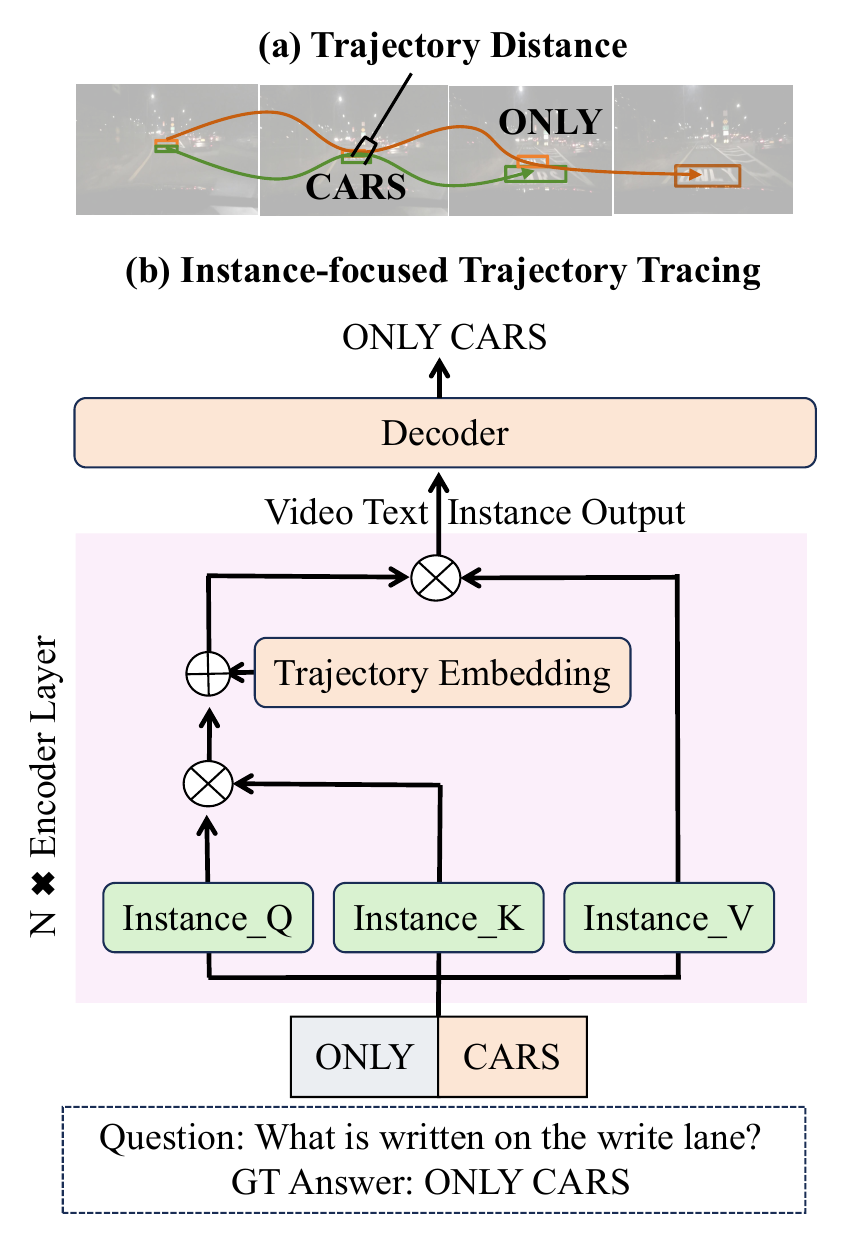}
  \caption{
  (a) An example for trajectory distance. The intermediate temporal overlap between the instance ``CARS" and the instance ``ONLY" occurs in the second frame. Within this frame, the trajectory distance is computed as \cref{equ:equation2}.
  (b) Illustration of instance-focused trajectory tracing. 
}
    \label{fig:figure3}
\end{figure}

As discussed in the \cref{section:3.1}, we have successfully extracted multiple instances, including their textual contents and spatio-temporal trajectories. Following general Video TextVQA methods \cite{tea,M4-ViteVQA}, we also employ a T5-like encoder-decoder Transformer architecture to encode instances and generate answers. Unlike prior Video TextVQA methods that implicitly establish the spatio-temporal relationships between video text within sampled frames, our proposed module considers explicit trajectory interactions for each instance across the entire video sequence. These interactions are essential for comprehending the video, as they provide a more direct and reasonable spatio-temporal representation of the instances. As illustrated in \cref{fig:figure3}, we propose instance trajectory-aware attention mechanism integrated into the encoder layers and retain the vanilla Transformer-based decoder to answer the given question.

\begin{table*}[t]
\caption{Accuracy and ANLS comparison with the latest methods on the M4-ViteVQA \textit{Task1Split1} and RoadTextVQA. The best results are highlighted in bold and the second-best results are \underline{underlined}. {*} indicates the reproduced results using the open-source code.}
\centering
\begin{tabular}{ccccccc}
\toprule
\multirow{3}{*}{Method}                              & \multicolumn{4}{c}{M4-ViteVQA \textit{Task1Split1}}                                     & \multicolumn{2}{c}{RoadTextVQA} \\ \cmidrule(l){2-5}  \cmidrule(l){6-7} 
                                  & \multicolumn{2}{c}{Validation}  & \multicolumn{2}{c}{Test}                       & \multicolumn{2}{c}{Validation}  \\ \cmidrule(l){2-5}  \cmidrule(l){6-7} 
                                  & Acc.(\%)       & ANLS(\%)       & Acc.(\%) & ANLS(\%)                            & Acc.(\%)       & ANLS(\%)       \\ \midrule
\multicolumn{1}{c|}{JuskAsk \cite{Juskask}}      & 10.81          & 15.40           & 10.05    & \multicolumn{1}{c|}{14.10}           & --           & --           \\
\multicolumn{1}{c|}{All-in-one-B \cite{allinone}} & 11.47          & 15.30           & 10.87    & \multicolumn{1}{c|}{14.80}           & --           & --           \\
\multicolumn{1}{c|}{GIT \cite{GIT}}          & --           & --           & --     & \multicolumn{1}{c|}{--}           & 29.58          & 35.23          \\
\multicolumn{1}{c|}{SINGULARITY \cite{Singularity}}  & --           & --           & --     & \multicolumn{1}{c|}{--}           & 24.62          & 30.79          \\ \midrule
\multicolumn{1}{c|}{Video-LLaVA{*} \cite{videollava} }  & 15.82          & 17.77          & 15.43    & \multicolumn{1}{c|}{17.15}          & 30.82          & 40.92          \\
\multicolumn{1}{c|}{VideoLLaMaA2{*} \cite{videollama2} } & 20.04          & 21.73          & 20.76    & \multicolumn{1}{c|}{23.55}          & 25.11          & 36.53          \\
\multicolumn{1}{c|}{NVILA{*} \cite{nvila} }        & \underline{37.89}          & \textbf{47.67} & \underline{37.73}    & \multicolumn{1}{c|}{\underline{47.23}} & \underline{49.98}          & \underline{57.22 }          \\
\multicolumn{1}{c|}{Qwen2-VL{*} \cite{Qwen2-VL} }     & 36.77           & 46.56           & 35.22     & \multicolumn{1}{c|}{45.84}           & 47.23           & 55.34           \\ \midrule
\multicolumn{1}{c|}{T5-ViteVQA \cite{M4-ViteVQA} }   & 23.17          & 30.10          & 22.17    & \multicolumn{1}{c|}{29.10}          & --           & --           \\
\multicolumn{1}{c|}{TEA-Base \cite{tea} }     & 34.45          & 42.91          & 31.70    & \multicolumn{1}{c|}{40.24}          & 44.43          & 51.69          \\
\multicolumn{1}{c|}{TEA-Large \cite{tea} }    & 37.49          & 46.38          & 34.78    & \multicolumn{1}{c|}{43.71}          & 48.14          & 54.85          \\
\multicolumn{1}{c|}{GAT-Base}    & 35.31          & 44.64          & 35.56    & \multicolumn{1}{c|}{45.21}          & 46.54          & 53.78          \\
\multicolumn{1}{c|}{GAT-Large}   & \textbf{38.01} & \underline{47.53}          & \textbf{38.30 }   & \multicolumn{1}{c|}{\textbf{48.23}}          & \textbf{50.23} & \textbf{58.12} \\ \bottomrule
\end{tabular}

\end{table*}

\subsubsection{Instance Trajectory-aware Attention}
As illustrated in \cref{fig:figure3} (b), within the encoder layers, we devise an instance trajectory-aware attention mechanism to further enhance the explicit interaction between instances. Given the spatio-temporal trajectories of two instances $i$ and $j$, we derive their trajectory distances by considering both their relative spatial relationships and temporal intersections. Specifically, we identify the central frame within the intersection as the most representative interactive video frame $f_t$. The trajectory position is determined by the relative spatial relationship between the two instances at the frame $f_t$:
\begin{equation}
    trajPos = [x_{f_t}^{i} - x_{f_t}^{j}; y_{f_t}^{i} - y_{f_t}^{j}],
\end{equation} 
where $x_{f_t}^{i}$ and $y_{f_t}^{i}$ represent the normalized horizontal and vertical coordinates of instance $i$ at the frame $f_t$, respectively.

Subsequently, the trajectory embedding is computed based on the sine and cosine function \cite{attention}:
\begin{equation}
    trajEmbed_{ij} = [f^{sin}(x_{f_t}^{i} - x_{f_t}^{j}); f^{cos}(y_{f_t}^{i} - y_{f_t}^{j})].
\label{equ:equation2}
\end{equation}    

Within the instance trajectory-aware attention, we incorporate trajectory embedding as an attention bias to explicitly leverage trajectory information, thereby generating the instance output, as shown in the following equation:
\begin{gather}
    \alpha_{i,j} = \text{softmax}(Inst_{Q} \cdot Inst_{K}^T + \text{trajEmbed}_{ij}), \\
    Inst_{out} = \alpha_{i,j} \cdot Inst_{v}, 
\end{gather}
where $Inst_{K}$, $Inst_{Q}$, $Inst_{v}$ denote the key, query, and value for the instance in the attention mechanism.

\subsubsection{Auto-regressive Prediction}
Following previous Video TextVQA methods, we initialize the GAT with the language model and train it using cross-entropy loss. In cases where multiple answers exist for a given question in the dataset, the task is formulated as a multi-label classification problem. The training loss is defined as follows:
\begin{equation}
L_{ce} = - \left( y_{gt} \log(y_{pred}) + (1 - y_{gt}) \log(1 - y_{pred}) \right)
\end{equation}
where $y_{pred}$ is the prediction and $y_{gt}$ is the ground-truth target.

\section{Experiments}

\subsection{Experimental Settings}
\subsubsection{Datasets and Evaluation Metrics}
Our method is evaluated on two public Video TextVQA datasets: M4-ViteVQA \cite{M4-ViteVQA} and RoadTextVQA \cite{roadtextvqa}. The M4-ViteVQA dataset consists of 8,511 video clips and 24,123 question-answer pairs from nine YouTube categories (e.g., shopping, traveling, driving). It defines two tasks: Task 1 involves training and testing across all categories, with two splits for robustness evaluation: \textit{Task1Split1} (vanilla testing) and \textit{Task1Split2} (generalization testing with content variation). Task 2 trains on seven categories and tests on the remaining two. The RoadTextVQA dataset focuses on driving assistance, with 3,222 videos and 10,500 question-answer pairs from road signs and driving footage.
For these datasets, we utilize the VQA Accuracy (Acc.)  metric \cite{biten2019scene} and the average normalized Levenshtein similarity (ANLS) \cite{biten2019scene} to evaluate the model performance.

\subsubsection{Implementation Details}
\begin{table*}[]
\centering
\caption{Accuracy and ANLS comparison with the latest methods on the M4-ViteVQA \textit{Task1Split2} and \textit{Task2}. The best results are highlighted in bold. {*} indicates the reproduced results using the open-source code.}
\begin{tabular}{ccccccccc}
\toprule
\multirow{3}{*}{Method}                             & \multicolumn{4}{c}{M4-ViteVQA \textit{Task1Split2}}                             & \multicolumn{4}{c}{M4-ViteVQA \textit{Task2}}                      \\ \cmidrule(l){2-5}  \cmidrule(l){6-9}
                                  & \multicolumn{2}{c}{Validation} & \multicolumn{2}{c}{Test}              & \multicolumn{2}{c}{Validation} & \multicolumn{2}{c}{Test} \\ \cmidrule(l){2-3}  \cmidrule(l){4-5} \cmidrule(l){6-7} \cmidrule(l){8-9} 
                                  & Acc.(\%)       & ANLS(\%)      & Acc.(\%) & ANLS(\%)                   & Acc.(\%)       & ANLS(\%)      & Acc.(\%)    & ANLS(\%)   \\ \midrule
\multicolumn{1}{c|}{JuskAsk \cite{Juskask} }      & 7.16           & 10.00         & 5.47     & \multicolumn{1}{c|}{8.60}  & 4.86           & 6.70          & 3.60        & 6.70       \\
\multicolumn{1}{c|}{All-in-one-B \cite{allinone} } & 6.85           & 9.20          & 5.66     & \multicolumn{1}{c|}{7.80}  & 4.20           & 5.00          & 3.28        & 4.60       \\ \midrule
\multicolumn{1}{c|}{Video-LLaVA{*} \cite{videollava} }  & 13.14          & 14.29         & 11.19    & \multicolumn{1}{c|}{12.02} & 10.89          & 13.23         & 9.38        & 11.80      \\
\multicolumn{1}{c|}{VideoLLaMaA2{*} \cite{videollama2} } & 18.30          & 19.63         & 18.33    & \multicolumn{1}{c|}{20.45} & 19.68          & 23.62         & 16.54       & 21.80      \\
\multicolumn{1}{c|}{NVILA{*} \cite{nvila} }        & 30.25          & 40.58         & 30.10    & \multicolumn{1}{c|}{41.52} & 23.79          & 32.89         & 22.89       & 30.34      \\
\multicolumn{1}{c|}{Qwen2-VL{*} \cite{Qwen2-VL} }     & 28.55          & 39.34         & 27.25    & \multicolumn{1}{c|}{38.45} & 22.95          & 32.65         & 21.23       & 28.79      \\ \midrule
\multicolumn{1}{c|}{T5-ViteVQA \cite{M4-ViteVQA}}   & 17.59          & 23.10          & 16.68          & \multicolumn{1}{c|}{23.80}          & 12.30          & 16.10          & 9.29           & 13.60          \\
\multicolumn{1}{c|}{TEA-Base \cite{tea} }     & 26.66          & 36.61         & 26.29    & \multicolumn{1}{c|}{36.00} & 20.73          & 28.18         & 17.28       & 26.03      \\
\multicolumn{1}{c|}{TEA-Large \cite{tea}  }    & 28.27          & 36.32         & 28.43    & \multicolumn{1}{c|}{38.13} & 22.83          & 30.21         & 18.83       & 28.90      \\
\multicolumn{1}{c|}{GAT-Base}    & 29.07          & 39.26         & 29.77    & \multicolumn{1}{c|}{40.71} & 21.65          & 30.88         & 21.65       & 29.83      \\
\multicolumn{1}{c|}{GAT-Large}   & \textbf{31.35}          & \textbf{41.33}         & \textbf{30.90 }   & \multicolumn{1}{c|}{\textbf{41.81}} & \textbf{24.54}          & \textbf{33.30}         & \textbf{22.13}       & \textbf{30.75}      \\ \bottomrule
\end{tabular}

\label{tab:table2}
\end{table*}

In the experimental setup, following previous Video TextVQA methods, we utilize pretrained language models as the foundational backbone for processing video text as well as reasoning answers. To be specific, based on the pretrained T5 \cite{t5}, we build two models with different capacities, namely GAT-Base and GAT-Large, which have 12+12 and 24+24 encoder-decoder layers, respectively. 

\subsection{Comparison with State-of-the-art Methods}
On the M4-ViteVQA and RoadTextVQA benchmarks, we evaluate GAT and compare it with existing approaches. It is evident that GAT-Large surpasses various types of current methods. (1) \textbf{Video-language Pretraining Methods.} These methods \cite{allinone, GIT, Juskask,Singularity} underperform on the Video TextVQA task due to their inability to read small video text. AII-in-one-B \cite{allinone} achieves an accuracy of 11.47\% on the M4-ViteVQA \textit{Task1Split1} validation set, and GIT \cite{GIT} reaches 29.58\% in accuracy on the RoadTextVQA validation set. (2) \textbf{Video-LLMs.} Video-LLMs \cite{videollama2, videollava, nvila, Qwen2-VL} integrate a visual encoder with the large language model \cite{llama2} by pretraining on a large-scale video-text or image-text data. To adapt these methods for Video TextVQA, we apply LoRA \cite{lora} to fine-tune on the corresponding dataset. Generally, Video-LLMs employ the common frame sampling strategy used in video understanding, which suffers from incomplete, blurred, or repetitive video text within sampled frames. Therefore, ignoring the dynamic nature of video text leads to suboptimal performance of Video-LLMs. For instance, while NVILA \cite{nvila} enhances both spatial and temporal resolutions, it underperforms GAT-Large by 0.57\% in terms of accuracy on the M4-ViteVQA \textit{Task1Split1} test set. (3) \textbf{Video TextVQA Methods.} Compared with state-of-the-art Video TextVQA methods, TEA-Large \cite{tea}, GAT-Large achieves higher performance by a margin of 3.52\% in accuracy on the M4-ViteVQA \textit{Task1Split1} test set. Although the same two-stage paradigm is adopted, we believe our proposed instance-oriented framework can better bridge the gap between the two stages.

\subsection{Analysis of Generalization}

In this section, we evaluate the generalization performance of our proposed method. Generally, most VideoQA methods are prone to overfit specific training sets, achieving high accuracy on trained video categories but exhibiting poor performance on others \cite{videoqa}. However, in practical applications, it is important to generalize model capabilities across different video domains. Following general experimental settings, the M4-ViteVQA dataset \cite{M4-ViteVQA} is split into two tasks to evaluate generalization capabilities. From results in \cref{tab:table2}, GAT achieves consistent improvements in accuracy and ANLS compared to both video-language pretraining methods \cite{Juskask, allinone} and Video-LLMs \cite{videollama2, videollava, nvila, Qwen2-VL}. We attribute this performance to the proposed instance-oriented framework. GAT reads and understands instances across entire video frames, which are shard across various scenarios, including even in unseen video categories.

\subsection{Ablation Studies}
\subsubsection{Overall Results}

\begin{table}[]
\caption{Overall ablation results on the M4-ViteVQA \textit{Task1Split1} validation set. Gather and Trace represent context-aggregated instance gathering and instance-focused trajectory tracing, respectively. Random and Max are two heuristic methods for representing the textual content of each instance. Spatial refers to the sole consideration of spatial interaction, without accounting for the temporal factor.}
\begin{tabular}{ccccc}

\toprule
\multirow{2}{*}{\#} & \multirow{2}{*}{Gather} & \multirow{2}{*}{Trace}  & \multicolumn{2}{c}{M4-ViteVQA} \\ \cmidrule{4-5} 
                    &                              &                              & Acc.(\%)       & ANLS(\%)      \\ \midrule
a                   & Random                       & \multicolumn{1}{c|}{\checkmark}     & 32.36          & 42.84         \\
b                   & Max                          & \multicolumn{1}{c|}{\checkmark}     & 33.79          & 43.31         \\
c                   & \checkmark                          & \multicolumn{1}{c|}{$\times$}       & 26.89          & 35.85         \\
d                   & \checkmark                          & \multicolumn{1}{c|}{Spatial} & 33.68          & 43.11         \\
e                   & \checkmark                          & \multicolumn{1}{c|}{\checkmark}     & \textbf{35.31}          & \textbf{44.64}         \\ \bottomrule
\end{tabular}

\label{tab:table3}
\end{table}

\begin{table}[]
\caption{Ablation experiments about context-aggregated instance gathering module on ICDAR15 dataset. }
\begin{tabular}{clll}
\toprule
Method                                & \multicolumn{1}{c}{MOTA} & \multicolumn{1}{c}{IDF1} & \multicolumn{1}{c}{MOTP} \\ \midrule
\multicolumn{1}{c|}{GoMatching \cite{gomatching} }       & 72.04                    & 80.11                    & 78.53                    \\
\multicolumn{1}{c|}{GoMatching \cite{gomatching} + Gather} & \textbf{72.82}                    & \textbf{80.21}                    & \textbf{78.55}                    \\ \bottomrule
\end{tabular}

\label{tab:table4}
\end{table}

To assess the impact of GAT's key components, including context-aggregated instance gathering module and instance-focused trajectory tracing module, we perform ablation experiments on the M4-ViteVQA \textit{Task1Split1} validation set \cite{M4-ViteVQA}. Each module is incrementally integrated to evaluate its effect on model performance. The comparison results are detailed in \cref{tab:table3}, while the hyperparameters for each module are elaborated in the subsequent sections. 

To provide the unified textual representation for each instance, context-aggregated instance gathering module integrates the rich context information of instances throughout the entire video sequence. We perform three experimental settings (\#a, \#b, and \#e) to demonstrate the effectiveness of our proposed module. In row \#a, we randomly select an individual frame recognition result for the instance to denote its unified textual representation. In row \#b, we select the most frequent recognition result for the instance across the video to fulfill the same purpose. Row \#e presents our proposed module. From the results, our proposed module achieves a significant improvement of 2.95\% and 1.52\% in accuracy over the two heuristic methods (\#a, \#b), respectively. This improvement is attributed to the aggregation of rich multimodal features throughout the video sequence, including visual features, layout characteristics, and textual contents.

Additionally, we also explore the significance of instance-focused trajectory tracing module, which establishes explicit spatio-temporal interactions between video text instances. Comparisons between rows \#c and \#e indicate that removing our proposed module results in a performance decrease of 8.79 \% in ANLS. It reveals that spatio-temporal interactions within the instances are crucial for video text understanding. To further investigate the temporal factor within trajectories, we conduct experiments without and with considering the temporal intersection between instances. We note that considering the temporal factor results in a significant increase in performance from 33.68\% (row \#d) to 35.31\% (row \#e) in accuracy.

\subsubsection{Evaluation of Reading Video Text}
To adapt the spotted results from the VTS method for the Video TextVQA task, we introduce the context-aggregated instance gathering module in \cref{section:3.1}. Building upon GoMatching \cite{gomatching}, this module adopts an instance-oriented framework to perceive video text and generates a unified textual representation for each instance. As demonstrated in \cref{tab:table4}, we surprisingly note that the instance-oriented perception design not only boosts performance on the Video TextVQA task but also improves results on the VTS task. Specifically, compared with GoMatching \cite{gomatching}, this module obtains a 0.78\% improvement in MOTA on the ICDAR15 dataset \cite{icdar15}. We attribute this improvement to the consideration of the dynamic nature of video text during recognition in context-aggregated instance gathering module, unlike previous VTS methods that rely solely on visual appearance within individual frames.

\subsection{Analysis of Efficiency}
\subsubsection{Evaluation of Model Complexity and Inference Speed}
To evaluate the efficiency of the proposed instance-oriented framework, we compare the model complexity and inference speed on the M4-ViteVQA validation set \cite{M4-ViteVQA} with a single A6000 GPU. As shown in \cref{tab:table6}, GAT demonstrates superior inference speed compared to both Video-LLMs \cite{Qwen2-VL, nvila} and Video TextVQA methods \cite{tea}. Specifically, Video-LLMs, such as Qwen2-VL \cite{Qwen2-VL} and NVILA \cite{nvila}, which are built on the large-scale language model (7B parameters), exhibit slower inference speed than Video TextVQA methods. Furthermore, we compare TEA-Base \cite{tea} and GAT-Base$\dagger$ (a variant of GAT-Base without context-aggregated instance gathering module) that leverage the same language model \cite{t5} as the reasoning backbone. TEA-Base feeds redundant video text within sampled frames and generates video text-aware clues to assist in answering questions, but this incurs computational overhead and additional 1.2B model parameters. By reducing the token lengths of redundant video text, GAT-Base significantly improves inference speed compared to GAT-Base†. 

\begin{table}[]
\caption{
Efficiency comparison with the latest methods on the M4-ViteVQA \textit{Task1Split1} validation set.}
\centering
\begin{tabular}{ccc}
\toprule
Method             & Param. Size                & FPS  \\ \midrule
Qwen2-VL{*} \cite{Qwen2-VL}           & \multicolumn{1}{c|}{8B}   & 0.47 \\
NVILA{*} \cite{nvila}              & \multicolumn{1}{c|}{8B}   & 0.54 \\ \midrule
TEA-Base{*} \cite{tea}           & \multicolumn{1}{c|}{1.4B} & 1.09 \\
GAT-Base$\dagger$ & \multicolumn{1}{c|}{230M} & 2.41 \\
GAT-Base          & \multicolumn{1}{c|}{230M} & \textbf{5.46} \\ \midrule
\end{tabular}

\label{tab:table6}
\end{table}

\subsubsection{Token Length Analysis}
\label{sec:section4.5.2}
\begin{figure}[t]
    \centering  
  \includegraphics[width=0.3\textwidth]{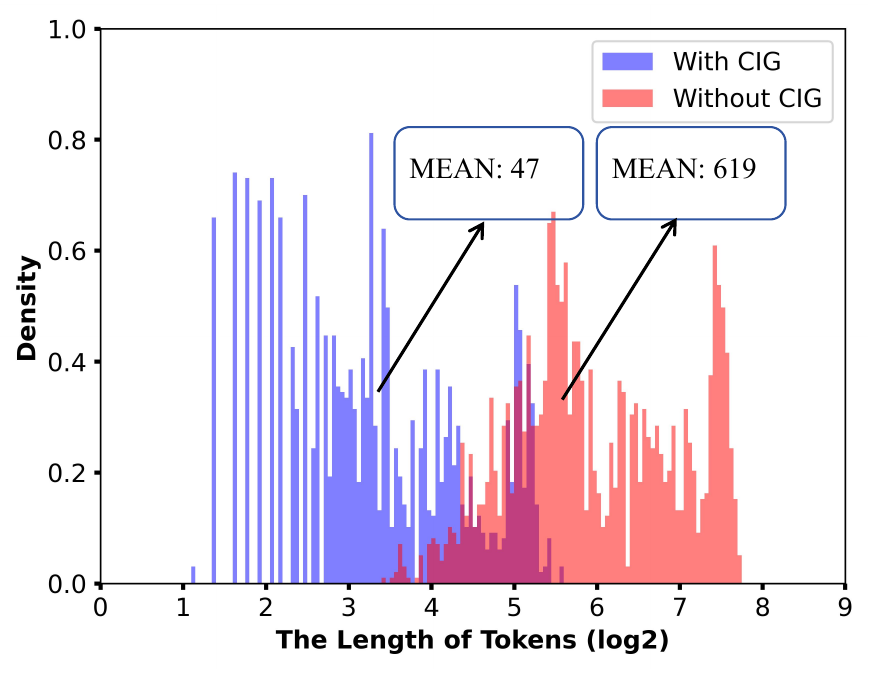}
  \caption{
  A histogram representing the token length of with and without the Context-aggregated Instance Gathering (CIG) module, respectively.
}
    \label{fig:figure4}
\end{figure}

To further investigate the reason why our method exhibits advantages in efficiency, we analyze the input token length on the M4-ViteVQA \textit{Task1Split1} validation set \cite{M4-ViteVQA}, as shown in \cref{fig:figure4}. The average token length with and without context-aggregated instance gathering module is 47 and 619, respectively. It means that our proposed method feeds non-redundant instances into the generative model, thus achieving significant improvement in inference speed while maintaining high accuracy.

\subsection{Qualitative Analysis}

\begin{figure}[]
    \centering  
  
  \includegraphics[width=0.45\textwidth]{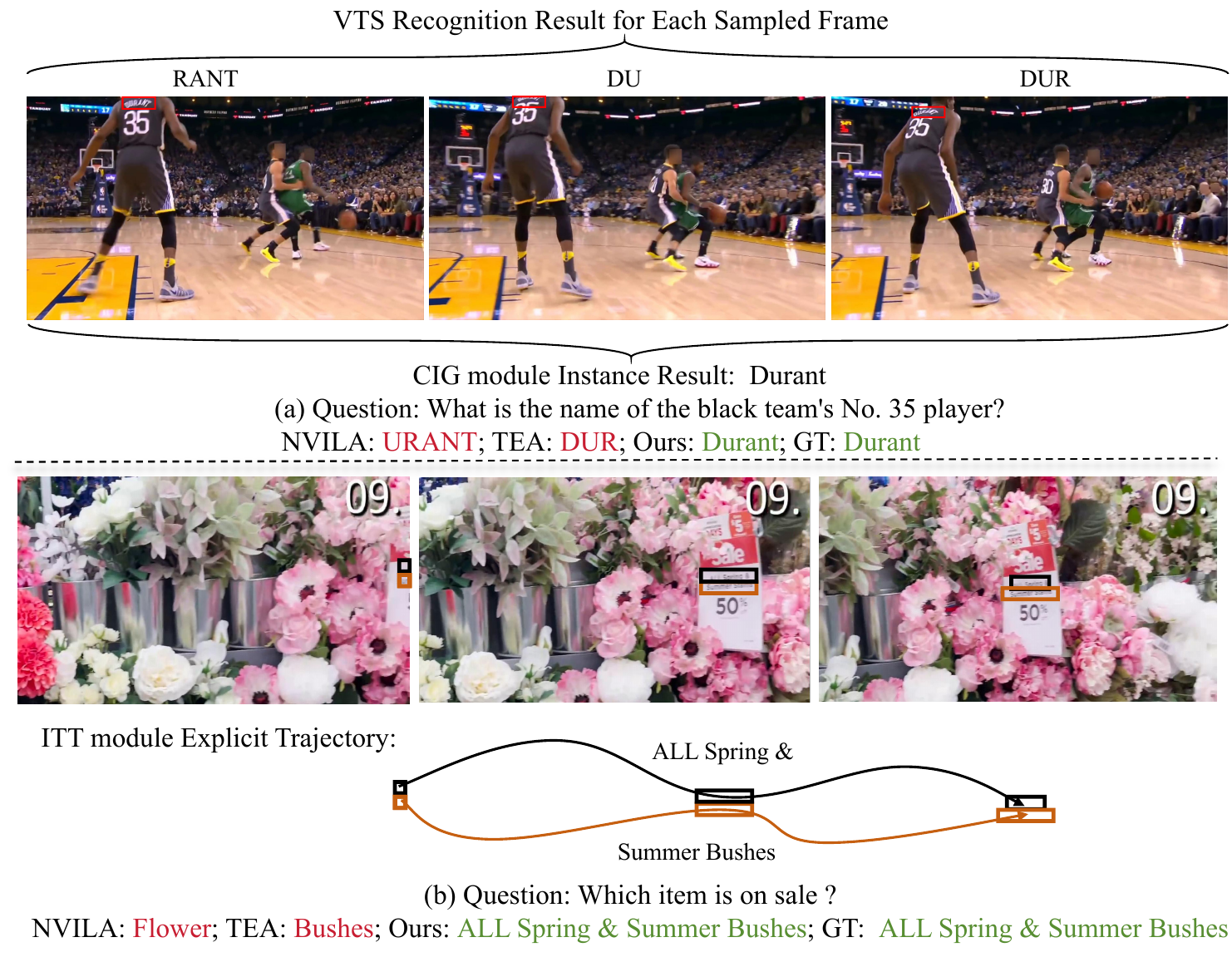}
  \caption{
  Qualitative examples. CIG and IIT, in the figure, denote the Context-aggregated Instance Gathering module and the Instance-focused Trajectory Tracing module, respectively. The red, black and brown boxes indicate the regions related to the question, respectively. Best viewed in zoom.
}
    \label{fig:figure5}
\end{figure}

As illustrated in \cref{fig:figure5}, two representative qualitative examples are provided to demonstrate the superiority of our proposed. Due to the dynamic nature of video text, there are many low-quality text across sequential video frames. Conventional frame-sampling strategy employed in Video-LLMs and Video TextVQA is prone to failure in such scenarios, while GAT can effectively address these challenges. In \cref{fig:figure5} (a), the instance ``Durant'' appears incomplete, blurred, and redundant across multiple frames. GAT integrates the rich context information of the instance throughout the video into a unified representation and leverages instance trajectory-aware mechanism to enhance the accuracy of answering. In \cref{fig:figure5} (b), the instance ``ALL Spring \&'' and the instance ``Summer Bushes'' appear in consecutive frames and form corresponding trajectory. The instance-focused trajectory tracing module explicitly establishes the spatio-temporal relationships between these instances and predicts the correct answer ``ALL Spring \& Summer Bushes''. More visualization cases can be found in the appendix.

\section{Conclusion}
In this paper, we focus on the limitations of frame-level framework employed in previous Video TextVQA methods and rethink how to deal with dynamic video text in videos. To address issues in existing work, we introduce the video text instance-oriented framework, referred to as GAT (Gather and Trace). By aggregating multimodal context features across the entire video sequence, GAT accurately reads the textual content of video text instance. We also design an instance trajectory tracing module to enhance the relationship modeling and facilitate interactions with the questions. Experimental results show that GAT exhibits advantages in both accuracy and efficiency. We expect this framework will have a potential impact on Video TextVQA research in the future.

\section{Acknowledgments}
This work is supported by the National Natural Science Foundation of China (Grant NO 62376266 and 62406318).

\bibliographystyle{ACM-Reference-Format}
\bibliography{sample-base}


\end{document}